\pdfoutput=1
\documentclass[sigconf]{acmart}
\renewcommand\footnotetextcopyrightpermission[1]{}
\usepackage{graphicx}
\usepackage{subfigure}
\usepackage{subcaption}
\usepackage{booktabs} % for professional tables
\usepackage{color}
\usepackage{tabularray}
\usepackage{xcolor}
\usepackage{amsmath}
\usepackage{dsfont}
\usepackage{multirow}
\usepackage{diagbox}
\usepackage{colortbl}
\usepackage{mathtools}
\usepackage{amsthm}
\usepackage{multirow}
\newcommand{\ie}{\emph{i.e.}}
\newcommand{\eg}{\emph{e.g.}}

\AtBeginDocument{%
  \providecommand\BibTeX{{%
    \normalfont B\kern-0.5em{\scshape i\kern-0.25em b}\kern-0.8em\TeX}}}

\setcopyright{acmlicensed}
\copyrightyear{2024}
\acmYear{2024}
\acmDOI{XXXXXXX.XXXXXXX}

\acmConference[ACM MM]{}{28 October - 1 November,
  2024}{Melbourne, Australia}
\acmISBN{978-1-4503-XXXX-X/18/06}

\usepackage{dsfont}
\usepackage{bbm}
\begin{document}

%%
%% The "title" command has an optional parameter,
%% allowing the author to define a "short title" to be used in page headers.
\title{Exploring the Distinctiveness and Fidelity of the Descriptions Generated by Large Vision-Language Models}

\author{Yuhang Huang}
\authornote{Both authors contributed equally to this research.}
\email{huangyh723@seu.edu.cn}
\affiliation{%
  \institution{Southeast University}
  \city{Nanjing}
  \state{Jiangsu}
  \country{China}
}

\author{Zihan Wu}
\authornotemark[1]
\email{zihanwuseu@outlook.com}
\affiliation{%
  \institution{Southeast University}
  \city{Nanjing}
  \state{Jiangsu}
  \country{China}
}

\author{Chongyang Gao}
\email{chongyanggao2026@u.northwestern.edu}
\affiliation{%
  \institution{Northwestern University}
  \city{Evanston}
  \state{Illinois}
  \country{United States}
}

\author{Jiawei Peng}
\email{pengjiawei@seu.edu.cn}
\affiliation{%
  \institution{Southeast University}
  \city{Nanjing}
  \state{Jiangsu}
  \country{China}
}

\author{Xu Yang}
\authornote{Corresponding author.}
\email{xuyang_palm@seu.edu.cn}
\affiliation{%
  \institution{Southeast University}
  \city{Nanjing}
  \state{Jiangsu}
  \country{China}
}

%%
%% Keywords. The author(s) should pick words that accurately describe
%% the work being presented. Separate the keywords with commas.

%% A "teaser" image appears between the author and affiliation
%% information and the body of the document, and typically spans the
%% page.

% \received{20 February 2007}
% \received[revised]{12 March 2009}
% \received[accepted]{5 June 2009}

%%
%% This command processes the author and affiliation and title
%% information and builds the first part of the formatted document.
\begin{abstract}
% Recognizing the limitations of existing Large Vision-Language Models (LVLMs) in producing detailed, fine-grained textual descriptions, this study develops a novel evaluation framework employing distinctiveness and fidelity metrics. These metrics rigorously assess how models like Open-Flamingo, IDEFICS, MiniGPT-4, and LLaVA distinguish between visually similar objects and the accuracy with which their descriptions mirror the visual content. By addressing these challenges, the framework not only sheds light on the current shortcomings of LVLMs but also offers guidance for enhancing their performance. Additionally, this research innovates by adapting LVLMs for classification tasks, extending their use beyond traditional generative contexts. This approach leverages their generative nature to solve classification challenges effectively, offering a fresh perspective on the potential of LVLMs in new applications. The study also explores the variability in how different LVLMs handle fine-grained descriptions, providing valuable insights into their capabilities for interpreting and generating nuanced visual narratives. This comprehensive analysis advances our understanding of multimodal language models, contributing significantly to the development of more sophisticated and versatile AI tools.
Large Vision-Language Models (LVLMs) are gaining traction for their remarkable ability to process and integrate visual and textual data. Despite their popularity, the capacity of LVLMs to generate precise, fine-grained textual descriptions has not been fully explored. This study addresses this gap by focusing on \textit{distinctiveness} and \textit{fidelity}, assessing how models like Open-Flamingo, IDEFICS, and MiniGPT-4 can distinguish between similar objects and accurately describe visual features. We proposed the Textual Retrieval-Augmented Classification (TRAC) framework, which, by leveraging its generative capabilities, allows us to delve deeper into analyzing fine-grained visual description generation. This research provides valuable insights into the generation quality of LVLMs, enhancing the understanding of multimodal language models. Notably, MiniGPT-4 stands out for its better ability to generate fine-grained descriptions, outperforming the other two models in this aspect. 
The code is provided at \url{https://anonymous.4open.science/r/Explore_FGVDs-E277}.

\end{abstract}

% \begin{figure}[t]
%   \centering
%   \begin{subfigure}{\linewidth}
%     \centering
%     \includegraphics[width=\linewidth]{fig/icl_sl.pdf}
%     \caption{In-context Learning by Single Label.}
%     \label{fig:intro-a}
%   \end{subfigure}
%   \begin{subfigure}{\linewidth}
%     \centering
%     \includegraphics[width=\linewidth]{fig/icl_ld.pdf}
%     \caption{In-context Learning by LDE.}
%     \label{fig:intro-b}
%   \end{subfigure}
%   \begin{subfigure}{\linewidth}
%     \centering
%     \includegraphics[width=\linewidth]{fig/icl_vd.pdf}
%     \caption{In-context Learning by VDE.}
%     \label{fig:intro-c}
%   \end{subfigure}
%    \begin{subfigure}{\linewidth}
%     \centering
%     \includegraphics[width=\linewidth]{fig/LS.pdf}
%     \caption{Manipulation of Label Space.}
%     \label{fig:intro-d}
%   \end{subfigure}
%   \caption{(a) Traditional in-context learning with single labels may fail to capture the correct ground truth label. (b) Our Label Distribution Enhancement (LDE) addresses this by considering a broader label distribution. (c) Visual Description Enhancement (VDE) further refines the accurate representation by incorporating detailed visual descriptors. (d) The combined manipulation of the label space with LDE and VDE enables a more precise and comprehensive classification.}
%    \label{fig:intro}
% \end{figure}
\maketitle
\section{Introduction}
\label{sec:intro}
\begin{figure}[t]
  \centering
    \includegraphics[width=0.97\linewidth]{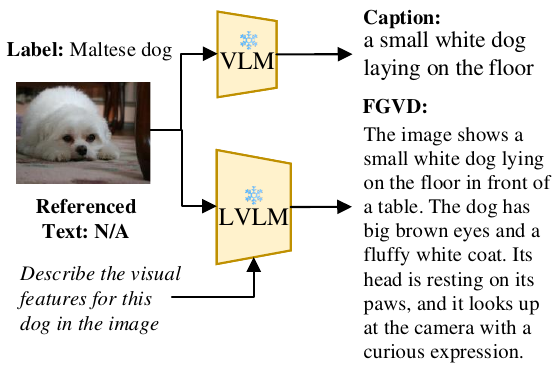}
    \caption{The caption produced by a smaller Vision Language Model (VLM) offers a broad overview of the image. In contrast, the fine-grained visual description (FGVD), generated by the Large Vision Language Model (LVLM) conditioned on both visual and linguistic cues, encompasses more nuanced details. }
    \label{fig:intro}
\end{figure}
In the evolving landscape of artificial intelligence, Vision-Language Models (VLMs) such as CLIP and BLIP~\cite{radford2021learning,li2022blip} have been instrumental in enhancing machine interpretation of visual content by aligning the visual and text embedding spaces through contrastive learning techniques. Building on this foundation and laying the groundwork for generating coherent, relevant captions across a diverse range of visual content~\cite{chang2018fine}, Large Vision-Language Models (LVLMs) like Open-Flamingo~\cite{awadalla2023openflamingo}, IDEFICS~\cite{laurenccon2024obelics}, and MiniGPT-4~\cite{zhu2023minigpt} have advanced the field further. These innovations have significantly improved machine capability to integrate visual content with textual descriptions, facilitating a broad spectrum of Visual-Language tasks, including Image Captioning(IC)~\cite{shao2023fine,huang2021image}. However, existing models focus on generative methods~\cite{zhang2024vision}, and the current solution for evaluating the quality of the generated text is relatively simplistic, making it difficult to effectively discern the quality differences between texts, particularly in terms of accurate and detailed descriptions of images. This limitation highlights a promising area for further research.

To address this gap, exploring the differences in detailed generation quality among LVLMs is required. However, generating accurate, fine-grained visual descriptions (FGVDs) and capturing the nuanced details essential~\cite{cho2022fine} continues to present significant challenges. Initially, these challenges are especially pronounced in fields that demand exceptional specificity, such as biodiversity for precise species identification~\cite{fabian2023multimodal}, and in technical domains that require differentiation between closely related yet distinct objects. Also, traditional captioning datasets such as Microsoft COCO~\cite{chen2015microsoft}, Flickr30k~\cite{plummer2015flickr30k}, and Visual Genome~\cite{krishna2017visual} with images containing a multitude of subjects and varied elements, often only provide broad overviews in their ground-truth descriptions. Specifically, their efficacy in producing fine-grained descriptions is significantly hampered by their inherent brevity, simplicity, and multi-target focus. Furthermore, traditional evaluation metrics like BLEU~\cite{papineni2002bleu}, CIDEr~\cite{vedantam2015cider}, and SPICE~\cite{anderson2016spice} are inadequate for assessing fine-grained descriptions because they rely heavily on these coarse ground-truth descriptions, which lacks the necessary granularity. 

These limitations have underscored a critical gap in evaluating the capabilities of LVLMs to deliver detailed descriptions. In response to the aforementioned issues, some researchers have proposed the CLIP score as an alternative to traditional n-gram-based metrics. Although it offers some advantages, the CLIP score still cannot capture fine-grained nuances because the contrastive learning process of the CLIP model still relies on images and coarse sentence pairs. Consequently, it remains an imperfect metric for evaluating the quality of detailed textual generation. This underscores the need to reevaluate methods and develop insights for fine-grained textual generation that overcome the constraints of traditional captioning datasets, calling for novel approaches to better understand the ability of  LVLM to produce high-quality visual descriptions.

In this paper, to tackle the issues outlined above, we propose a framework to analyze the quality of FGVDs generated by LVLMs from two perspectives. We focus on \emph{distinctiveness} because LVLMs effectively recognize the broad categories to which images belong. This aspect is explored from a fine-grained perspective to further understand and enhance their capabilities. This is crucial because of the comparative advantages of generative LVLMs over non-generative models like CLIP, which do not specifically tailor their outputs to address the complexities inherent in fine-grained datasets.
In contrast to earlier models that heavily relied on captioning datasets such as Microsoft COCO, current LVLMs demonstrate marked improvements in the granularity of their descriptions, which makes it urgent to evaluate the precision and relevance of generated descriptions with fine-grained datasets. To overcome these limitations, this study utilizes fine-grained classification datasets tailored to specific categories~\cite{chen2020say}, which are distinct from traditional captioning datasets. These resources make it possible to evaluate the extent to which LVLM-generated texts can effectively distinguish among categories at a granular level. We examine the ability of these models to generate descriptions that effectively distinguish among diverse categories by precisely identifying and articulating nuances between closely related visual subjects. However, 
the generated descriptions might inaccurately reflect the visual content, thereby necessitating the inclusion of \emph{fidelity}. Fidelity assesses from a more coarse-grained viewpoint to ensure that the generated text not only differentiates effectively but also remains accurate and faithful to the original images. This balanced approach—from distinctiveness to fidelity enhances our understanding of model performance in the intricate task of fine-grained text generation, emphasizing the importance of both granular detail and overall accuracy in the evaluation process.
% \begin{figure}[t]
%   \centering
%     \includegraphics[width=0.97\linewidth]{fig/intro.pdf}
%     \caption{The caption produced by a smaller Vision Language Model (VLM) offers a broad overview of the image. In contrast, the fine-grained visual description (FGVD), generated by the Large Vision Language Model (LVLM) conditioned on both visual and linguistic cues, encompasses more nuanced details. }
%     \label{fig:intro}
% \end{figure}

To evaluate \emph{distinctiveness}, our approach draws inspiration from the Retrieval-Augmented Generation (RAG) framework~\cite{lewis2020retrieval} within NLP. We have developed a method named "\textbf{T}extual \textbf{R}etrieval-\textbf{A}ugmented \textbf{C}lassification"\textbf{(TRAC)}, which utilizes classification-driven methods to explore how LVLMs distinguish unique attributes among closely related categories. First, a subset of generated descriptions is reserved as a reference corpus against which test descriptions are compared. The distinctiveness of these descriptions is then indirectly assessed by measuring the accuracy with which they are categorized under the correct labels, providing a way to gauge the uniqueness of the generated FGVDs. For \emph{fidelity}, It includes two approaches: firstly, applying the CLIP embedding similarity to assess the alignment between images and their textual outputs showcases the direct text-image relationship; secondly, employing the Stable Diffusion model to transform LVLM-generated FGVDs back into images, inspired by the concept of text as an effective cross-modal interface~\cite{wei2023diffusion}. This allows for fidelity evaluation through image-to-image comparison. Through the evaluation of distinctiveness and fidelity, this method facilitates a comprehensive analysis of the quality of fine-grained visual descriptions generated by LVLMs, thereby capturing their nuanced capabilities and performance from several critical perspectives.

By examining the nuances of generating detailed image descriptions with LVLMs from multiple perspectives, this study aims to deepen the understanding of how these models perform in creating fine-grained visual descriptions. This comprehensive analysis explores various aspects of description generation, highlighting the capabilities and limitations of LVLMs in this context. Contributions of this research include:
\begin{enumerate}

\item We are the first to evaluate the quality of LVLM-generated fine-grained visual descriptions through distinctiveness and fidelity, enriching multimodal language model research and identifying areas for system refinement.

\item We introduced a novel method Textual Retrieval-Augmented Classification (TRAC) that utilizes LVLMs for classification, inspired by the RAG framework. This approach addresses the unique challenges posed by the generative nature of LVLMs, unlike contrastive models like CLIP. It thereby broadens the tasks of LVLMs and provides new insights into their generations.

\item We determined the conditions under which LVLMs produce hallucinations in detailed FGVDs. By integrating high-quality GPT-4 descriptions with In-Context Learning, we significantly improved the ability of LVLMs to produce more detailed and accurate text.

\end{enumerate}
% \begin{itemize}
%     \item Our study is the first to assess the quality of textual descriptions generated by LVLMs through the lenses of distinctiveness and fidelity. This approach enriches existing research on multimodal language models and provides crucial insights for refining these systems. By identifying strengths and limitations, our analysis pinpoints areas for improvement, offering actionable insights that enhance the robustness and accuracy of LVLM.
%     \item We have developed a new method that leverages the generative nature of Large Vision-Language Models (LVLMs) for classification tasks, inspired by the Retrieval-Augmented Generation (RAG) framework. Unlike contrastively optimized models like CLIP, LVLMs present unique challenges for classification due to their generative focus. Our approach uses these generative features in combination with RAG principles to effectively address classification challenges, expanding the applications of LVLMs and providing fresh insights into their integration with classification frameworks for varied research and practical uses.
%     \item We identified the conditions under which LVLMs generate hallucinations when tasked with producing fine-grained distinguishing descriptions by analyzing Distinctiveness and Fidelity and providing insights into these phenomena. Additionally, we discovered that utilizing high-quality descriptions generated by GPT-4, combined with In-Context Learning, is able to enhance the model to generate finer-grained and more discriminative text.
% \end{itemize} 
\section{Related Works}
\label{sec:related}

%-------------------------------------------------------------------------
\subsection{Generating Visual Descriptions}
Image captioning (IC) is a pivotal task in computer vision that involves translating images into textual descriptions. Traditionally, numerous captioning models~\cite{vinyals2015show,anderson2018bottom,herdade2019image} have been developed, employing encoder-decoder architectures to achieve high-quality caption generation. In recent years, advancements have been made through the extensive training of models on large-scale image-text pair datasets and the integration of visual and linguistic modalities. Innovations such as ClipCap~\cite{mokady2021clipcap} and BLIP~\cite{li2022blip} have significantly improved the understanding of visual content, thereby enhancing the accuracy and contextual relevance of generated captions.

However, the captions generated by these models often provide only a broad overview of the image, lacking specificity. On the other hand, elaborate textual descriptions enhance comprehension of the visual cues within the text, benefiting tasks like text-to-image generation. The generation of these detailed descriptions, pioneered by these works~\cite{menon2022visual, maniparambil2023enhancing} using Large Language Models (LLMs) (\eg, GPT-3, GPT-4) has contributed to the improvement of zero-shot classification of CLIP model. Nonetheless, despite their detailed nature, these descriptions may still produce inaccuracies due to the absence of direct visual references, leading to descriptions that are not visually congruent with the actual features of the objects depicted. The emergence of Large and Vision Language Models (LVLMs) (\eg, Open-Flamingo~\cite{awadalla2023openflamingo}, MiniGPT-4~\cite{zhu2023minigpt}, IDEFICS\cite{laurenccon2024obelics}) offers the potential for generating more extended and detailed text by leveraging a wealth of external visual and linguistic knowledge. A recent study~\cite{fabian2023multimodal} has exploited this potential, using LVLMs to generate fine-grained visual descriptions for zero-shot wildlife animal species recognition. Similarly, our work employs LVLMs, prompted with both visual and linguistic cues (\ie, crafted questions), to produce rich and detailed visual descriptions at a more granular level. Diverging from prior efforts, our work aims to steer LVLMs to explore various perspectives of visual content via various crafted questions, which is vital for the subsequent analysis and evaluation of these generated descriptions.

\subsection{Vision Language Models (VLMs)}

Vision Language Models (VLMs) have become a central focus within artificial intelligence research. CLIP~\cite{radford2021learning} marks a pivotal development by utilizing a contrastive method and pre-training on 400 million image-caption pairs to effectively synchronize vision and language domains. Parallel advancements, such as ClipCap~\cite{mokady2021clipcap} and BLIP~\cite{li2022blip}, have improved the creation of visually coherent captions. The emergence of Large Language Models like GPT-3~\cite{brown2020language} and LLaMA~\cite{touvron2023llama} has facilitated the development of advanced LVLMs, which harness vast LLM knowledge to deeply understand images through sufficient alignment across modalities, employing diverse multimodal datasets for training. Models like Open-Flamingo~\cite{awadalla2023openflamingo} and IDEFICS~\cite{laurenccon2024obelics} achieve this through perceiver modules and cross-attention techniques, whereas MiniGPT-4~\cite{zhu2023minigpt} integrates a projection layer and Q-former with LLM, training on accessible datasets like LAION.

\subsection{Evaluation of Vision-Language Models (VLMs)}
The traditional evaluation of Vision-Language Models (VLMs) primarily focuses on their ability to understand and interpret visual content within images. For instance, the Image Captioning (IC) task, which assesses how accurately an image can be translated into text, typically employs metrics such as BLEU-4~\cite{papineni2002bleu} and CIDEr~\cite{vedantam2015cider} for measurement. However, with the increasing scale of VLMs, the scope of their evaluation has expanded to become more comprehensive and robust. Recent studies~\cite{liu2023mmbench, xu2023lvlm, yin2023survey} have contributed to the development of benchmarks for evaluating VLMs, covering aspects of visual perception, understanding, and reasoning. However, quantifying the evaluation criteria for VLMs in the absence of reference texts or answers remains a significant challenge. Although both automated evaluation using models like GPT-4 and manual evaluation by humans are theoretically viable approaches, they entail considerable costs. Consequently, focuses predominantly on evaluating the fine-grained visual descriptions generated by LVLMs, circumventing the need for GPT-4 and reference to ground-truth texts. To this end, we leverage publicly accessible models, namely CLIP~\cite{radford2021learning} and Stable Diffusion~\cite{rombach2021highresolution}, to assess the generated visual descriptions for distinctiveness and fidelity.

\section{Method}

\label{sec:method}
\begin{figure*}[h]
  \centering
    \includegraphics[width=0.80\linewidth]{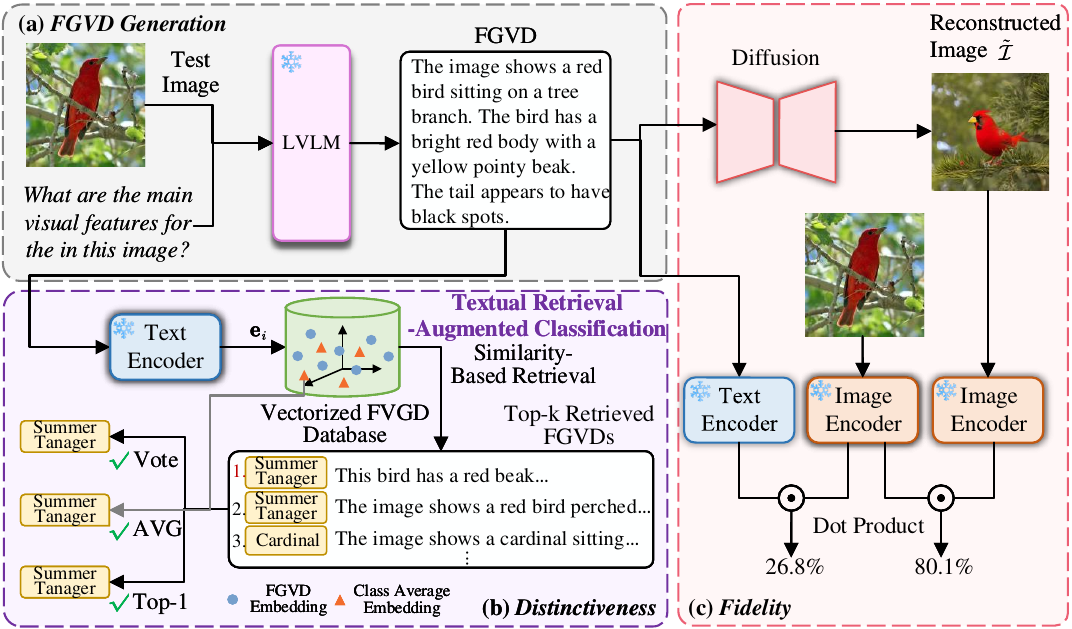}
    % \caption{This is a framework.}
    \caption{An overview of our framework for evaluating the quality of fine-grained visual descriptions (FGVDs) generated by Large Vision-Language Models (LVLMs). In the FGVD Generation phase (a), FGVDs are produced by conditioning on both visual and linguistic cues. Subsequently, we evaluate the quality of generated content in terms of its distinctiveness (b) and fidelity (c).}
    \label{fig:method}
\end{figure*}

This section outlines a systematic approach for assessing Large Vision-Language Models (LVLMs) in terms of their ability to produce detailed visual descriptions. In Section~\ref{sec:vdg}, the procedure for generating fine-grained visual descriptions (FGVDs) via LVLMs is elaborated. Following this, a dual-strategy framework is introduced for evaluating the quality of generated contents, focusing on distinctiveness (Section~\ref{sec:d}) and fidelity (Section~\ref{sec:f}). Fig.~\ref{fig:method} illustrates this evaluative framework, facilitating an in-depth exploration of the quality of FGVDs produced by LVLMs.

\subsection{Fine-Grained Visual Description Generation}
\label{sec:vdg}
The fine-grained visual descriptions are produced by the LVLMs (\eg, Open-Flamingo~\cite{awadalla2023openflamingo}) conditioned on both visual and textual cues. Contrary to the task of caption generation, which produces simple and broad descriptions, our objective requires the generation of more refined and nuanced textual descriptions. This demands the crafting of specific prompting questions designed to elicit detailed descriptions from the models. With textual prompt $\mathcal{P}$ and image $\mathcal{I}$, the process of fine-grained visual description generation can be formulated as follows:
\begin{equation}
    d=f_{\mathcal{M}}\left(\mathcal{I},\mathcal{P} \right)
\end{equation}
where $f_{\mathcal{M}}(\cdot)$ denotes the forward process of LVLM $\mathcal{M}$ and $d=\{\hat{w_1}, \dots, \hat{w_t}\}$ represents the sequence of generated words, decoded auto-regressively. For zero-shot generation, $\mathcal{P}$ represents a meticulously crafted prompting question $\mathcal{Q}$ aimed at eliciting either a salient or global description. For the in-context learning setting, $\mathcal{I}$ includes one query image and a set of demonstrated images. The textual prompts $\mathcal{P} $ are composed of a few interleaved demonstrated prompting questions $Q$ and descriptions $d$. 
%-------------------------------------------------------------------------
% The evaluation of fine-grained description generation encompasses two major modules: Eliciting fine-grained descriptions via LVLMs and Evaluating the generated descriptions. 

\subsection{Dual-Evaluation}
\label{subsec:eval}
\subsubsection{\textbf{Distinctiveness}}
\label{sec:d}

To evaluate the proficiency of LVLMs in differentiating among various categories, we implement a retrieval-based approach named \textbf{TRAC}. It involves utilizing the training set of a fine-grained dataset to construct a supporting set of fine-grained visual descriptions (FGVDs). These descriptions, generated by LVLMs for a predetermined set of images, are processed through a text encoder to extract embeddings, which are then stored for subsequent retrieval tasks.

\noindent \textbf{TRAC Process.} Let $D = \{d_1, d_2, \ldots, d_n\}$ denote the set of FGVD generated by the LVLMs, where $n$ is the number of images in the supporting set. Each FGVD $d_i$ corresponds to an image $\mathcal{I}_i$ and is associated with a label $y_i$. The embeddings extracted from these FGVD are represented as $ \mathcal{E} = \{\mathbf{e}_1, \mathbf{e}_2, \ldots, \mathbf{e}_n\}$, where $\mathbf{e}_i$ is the embedding of FGVD $d_i$.

For the testing set of the dataset, each image \(\mathcal{I}_{\text{test}}\) is similarly processed to obtain a textual description \(d_{\text{test}}\), which is then encoded into an embedding $\mathbf{e}$. This embedding is used to retrieve the most similar embeddings $\hat{\mathbf{e}}$ from \(\mathcal{E}\) by leveraging cosine similarity. The label \(y_{\text{R}}\) associated with the most similar embedding $\hat{\mathbf{e}}$ is then compared with the original label $y$ of \(\mathcal{I}_{\text{test}}\). The overall distinctiveness of the generated FGVD is quantified by calculating the accuracy across the testing set, defined as the proportion of instances where $\hat{y} = y$:
\begin{equation}
\text{Accuracy} = \frac{1}{|\mathcal{D}_{\text{test}}|} \sum_{\mathcal{I}_{\text{test}} \in \mathcal{D}_{\text{test}}} \mathds{1}(\hat{y} = y)
\end{equation}

where \(T\) represents the test set, and $\mathds{1}$ is the indicator function, equal to 1 when $\hat{y} = y$ and 0 otherwise. This accuracy metric serves as an indirect measure of the capability of LVLMs to generate FGVDs that are distinct enough to allow for accurate categorization and retrieval based on content similarity.

We also explored several approaches to assess the diversity of distinctiveness, initially concentrating on identifying the most similar description, referred to as $\text{TOP-1}$, as previously mentioned. Subsequently, we designed  two additional retrieval methods to expand our analysis:

\noindent \textbf{Average Embedding.}
For each category in the dataset $\mathcal{D}$, we calculate the average embedding of its diverse fine-grained visual descriptions. This process involves averaging the embeddings for all descriptions within a category, denoted as $\bar{\mathbf{e}}_c$. The embedding of a test description is then compared against $\bar{\mathbf{e}}_c$ to ascertain the closest match. This methodology capitalizes on the central tendency of embeddings within each category, reducing variance caused by individual description outliers and offering a condensed representation of the semantic space associated with each category. Additionally, this approach accentuates inter-class differences, sharpening the focus on distinguishing between categories to enhance classification precision.

\noindent \textbf{TOP-K Voting.}
This method selects the $\text{TOP-K}$ closest candidates rather than $\text{TOP-1}$ based on embedding similarity and determines the final category based on a majority vote among these candidates. This approach offers the advantage of mitigating the impact of outliers and ensures a more robust and democratic determination of the most representative category for each fine-grained visual description. Additionally, this method aims to enhance the discrimination of conditions within each class, ensuring finer resolution of intra-class variations.
\subsubsection{\textbf{Fidelity}}
\label{sec:f}

In addition to evaluating the distinctiveness of generated Fine-Grained Visual Descriptions (FGVDs), it is imperative to assess their fidelity, \ie, the extent to which these descriptions visually align with the original images. Relying solely on distinctiveness may lead to descriptions that, while unique, incorporate external and sometimes irrelevant information, thereby deviating significantly from the actual visual content of the source images. 
On the other hand, FGVDs serve as concise representations of visual content of an image, which, composed of thousands of pixels, contains much redundant information. Textual descriptions offer a means to compress this visual information efficiently. Determining the degree to which these descriptions preserve the original visual information is essential for ensuring the quality of the textual representations. To this end, we employ two strategies to investigate this fidelity:

\noindent\textbf{Image-Text Fidelity.} 
Drawing on methods akin to CLIP-score~\cite{hessel2021clipscore}, we use the embedding-similarity $s(\mathcal{I}, d)$ to measure the alignment between the input image $\mathcal{I}$ and the generated fine-grained visual description (FGVD) $d$.

\noindent\textbf{Image-Reconstructed Image Fidelity.} 
Fine-grained visual descriptions (FGVDs) serve not only as compact representations of images but also as effective cross-modal interface representations for multimodal tasks~\cite{wei2023diffusion}. Hence, we leverage Stable Diffusion~\cite{rombach2021highresolution} to manifest the semantic information hidden in the corresponding FGVD. The transformation is represented as:
\begin{equation}
    \tilde{\mathcal{I}} = \texttt{Diffusion}(\mathcal{I})
\end{equation}
Subsequently, we measure the similarity or distance $s(\mathcal{I}, \tilde{\mathcal{I}})$ between the origin image $\mathcal{I}$ and the reconstructed image $\tilde{\mathcal{I}}$, to quantify the preservation of visual content information via the compression of FGVD.

\section{Experiments}
\label{sec:experiments}
\begin{figure*}[htbp]
    \centering
    \includegraphics[width=1\linewidth]{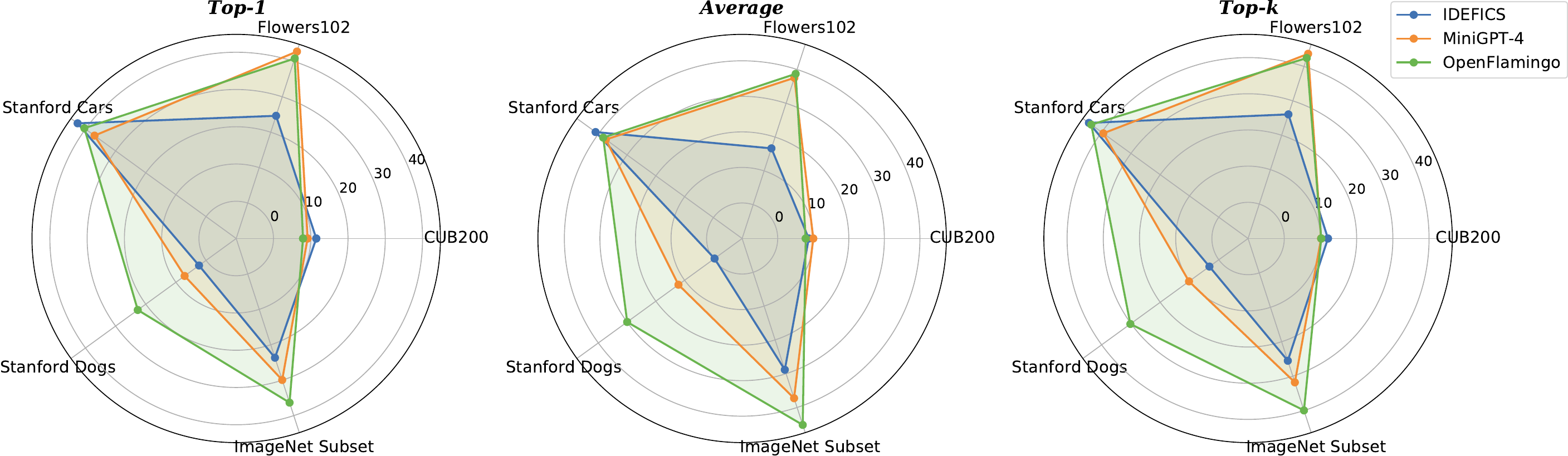}
    \caption{Results of LVLMs under different distinctness methods on five datasets.}
    \label{fig:overall_res1}
\end{figure*}

This section outlines our study implementations and experiments. In Section~\ref{subsec:dataset}, we describe the datasets used, including fine-grained classification sets and ImageNet1k, to evaluate LVLM performance across varied detail levels. We examine distinctiveness in Section~\ref{subsec:d} and fidelity in Section~\ref{subsec:fidelity}, culminating in a comprehensive analysis of hallucination issues in Section~\ref{subsec:quality}. Here, we assess the capabilities of LVLMs for generating detailed text descriptions, highlighting differences in quality, and offering insights into potential solutions for reducing hallucinations.

% \begin{table}
% \centering
% \caption{Dataset Information}
% \begin{tabular}{lccc}
% \toprule
% \textbf{Dataset} & \textbf{Classes} & \textbf{Support Set} & \textbf{Test Set} \\
% \midrule
% CUB200 & 200 & 5994 & 5794 \\
% Flowers102 & 102 & 4093 & 2463 \\
% Stanford Cars & 196 & 8144 & 8041 \\
% Stanford Dogs & 120 & 12000 & 8580 \\
% ImageNet Subset & 1000 & 10000 & 5000 \\
% \bottomrule
% \end{tabular}

% \label{tab:adjusted_test_datasets}
% \end{table}

% \usepackage{multirow}
% \usepackage{booktabs}

\subsection{Datasets and Implementation Details}
\label{subsec:dataset}

%-------------------------------------------------------------------------
% Datasets introduction

\noindent\textbf{Datasets.}
To assess the quality of fine-grained visual descriptions (FGVDs), our study encompasses a series of comprehensive experiments across five distinct image classification datasets, which span both fine-grained and more general object categories. We selected CUB-200~\cite{WelinderEtal2010}, Stanford Dogs~\cite{khosla2011novel}, Stanford Cars~\cite{krause20133d}, and Oxford 102 Flowers for fine-grained datasets. For the generic category, we utilized a specifically curated subset of ImageNet1k~\cite{deng2009imagenet}, comprising 10 images per category in the training set and 5 per category in the testing set. More detailed information about the dataset will be thoroughly presented in the Appendix. The chosen datasets, such as CUB200 and the Stanford\_Dogs dataset, are selected for their emphasis on fine-grained distinctions among closely related categories, requiring high visual specificity and intricacy in model predictions. This makes them ideal for assessing the quality of descriptions generated by Large Vision-Language Models (LVLMs), presenting a significant challenge in differentiating nuanced categories.

It is worth noting that for all the selected datasets, we apply a train/test split. Concretely, the training set is employed as a support set for retrieving FGVDs, and the testing set is reserved for evaluation.

%-------------------------------------------------------------------------
% Implementation details

\noindent\textbf{Implementation Details.}
In our experiments, we utilized three Large Vision-Language Models: Open-Flamingo~\cite{awadalla2023openflamingo} with the ViT-L/14 vision encoder, IDEFICS~\cite{laurenccon2024obelics}, and MiniGPT-4~\cite{zhu2023minigpt}. To ensure consistency across models, parameters were standardized, setting beam size to 3 and the length penalty to 1.0. To explore the impact of text length on performance, maximum generation length was treated as a variable, investigating its effect on the quality of generated descriptions. Notably, our experimental setup required no additional training or fine-tuning, utilizing a single NVIDIA RTX 3090 GPU with BF16 acceleration for all tests. For feature extraction from fine-grained visual descriptions (FGVDs), we chose the text encoder trained by CLIP, comparing it against alternatives like Sentence-BERT and the TF-IDF technique used in retrieval frameworks, as detailed in the Appendix. Additionally, our fidelity evaluations used the Stable Diffusion Model version 1.4~\cite{Rombach_2022_CVPR} to convert LVLM-generated FGVDs back into images.

%-------------------------------------------------------------------------
% Results and discussion

% \subsection{Results and Discussion}
\subsection{Distinctiveness Evaluation}
\label{subsec:d}

In this section, we present the experimental results for the distinctiveness aspect of our evaluation framework. We utilized a zero-shot approach and used the fixed prompt format for input into each Large Vision-Language Model, focusing on local visual features to emphasize the detailed aspects of an image. 
The prompt template was structured as follows:

\noindent \textit{"What are the main visual features for \{\textbf{Category}\} in this image?"}
Here, we provided only general categories to the prompt, such as "bird" for CUB200 and "dog" for Stanford Dogs, without offering any specific demonstrations. This zero-shot setting allows the LVLM to focus more on localized rather than global features of the images.

\noindent \textbf{Open-Flamingo Offers Superior Distinctiveness But Adds Extra Knowledge; MiniGPT-4 Remains Stable.} As shown in Fig. \ref{fig:overall_res1}, consistent trends were observed across the three methods used to evaluate model distinctiveness, with the Open-Flamingo model exhibiting superior performance on the aggregated five datasets. MiniGPT-4 displayed moderate performance, whereas IDEFICS underperformed. A detailed analysis of the generated results revealed that MiniGPT-4 and IDEFICS tend to produce well-structured formats, with descriptions closely matching the original images across most datasets. However, Open-Flamingo often deviated towards directly generating labels present in the images, avoiding visual descriptions, leading the large language model (LLM) to introduce additional, category-related information not strictly faithful to the original images.  This tendency, which enhances distinctiveness, especially in the Stanford Dogs dataset, is further detailed in the Appendix.
In contrast, \textbf{MiniGPT-4} adeptly avoided this issue, striving to describe the actual content depicted in the images. We speculate this is due to its training phases effectively aligning textual and visual features. This observation underscores the need to ensure a better alignment and emphasize the importance of incorporating fidelity in our evaluation.

\noindent \textbf{LVLMs Vary in Dataset-Specific Performance.} All three models exhibited poor performance in fine-grained differentiation within the CUB200 dataset, and they also struggled with the Stanford Dogs dataset in IDEFICS and MiniGPT-4. Conversely, the three LVLMs performed commendably on the Stanford Cars dataset. In contrast, animals like birds and dogs have higher intra-species similarity and subtle variations in features. Specifically, the challenge with animals could stem from their greater variation in poses, environments, and overlapping features, complicating the fine-grained classification task for LVLMs. We found that the primary descriptive features for birds focus on aspects such as feathers, beaks, and bellies, while cars, distinguished by logos and easily identifiable features of the car body such as the hood, windows, and tires, often require only brand recognition, bypassing the need to identify specific model years(\eg, "Acura TL sedan 2012", LVLMs can classify correctly only by describing it as Acura TL ). The average method demonstrated improvements for all models on the ImageNet dataset, suggesting that this method more effectively captures large category differences in datasets that are not fine-grained. Also, the trends we observed across the five datasets align closely with the classification scores produced by CLIP~\cite{radford2021learning, chen2023manipulating} on the same datasets. This parallelism underscores the efficacy of our approach, which solely relies on textual input for classification.

\begin{figure}[htbp]
    \centering
    \includegraphics[width=1\linewidth]{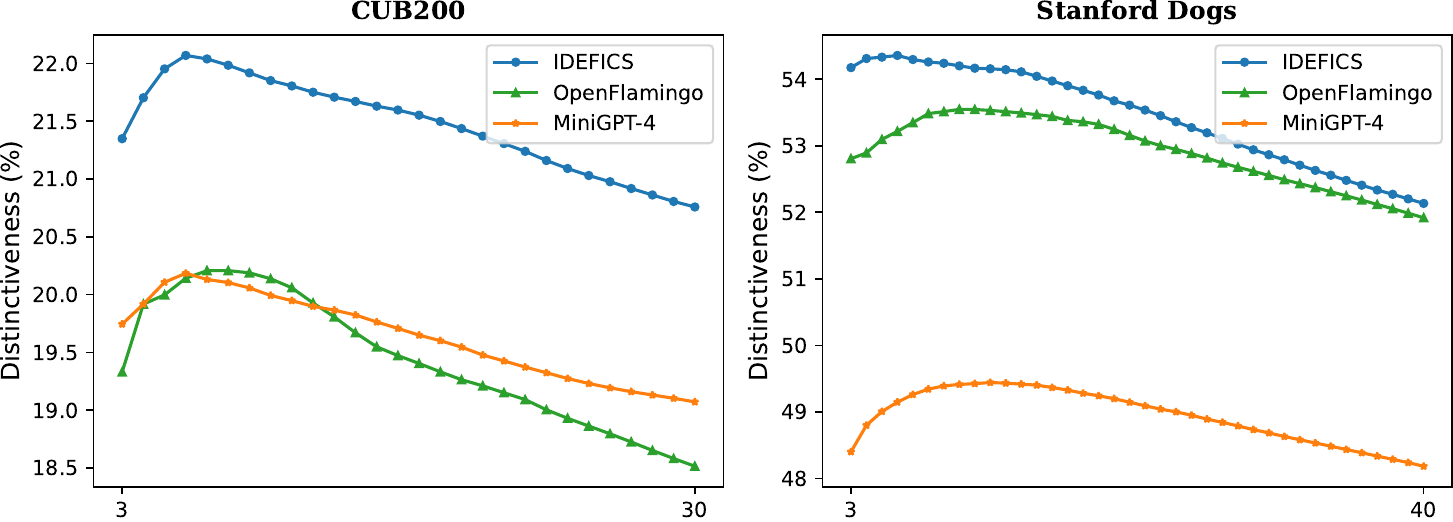}
    \caption{The distinctiveness results at different $k$-values}
    \label{fig:topk}
\end{figure}

\noindent \textbf{Top-k Exposes Poor Categorization Distinctiveness in LVLMs} For the top-k method, distinctiveness was assessed at various values of $k$, where $k \in (3, k_{max})$. $k_{max}$ was set based on the approximate number of images per class within each test dataset. (\eg, the CUB200 testing set averages 30 images per class, so $k$ ranged from 3 to 30) Results indicate, as shown in Fig. \ref{fig:topk}, that differentiation tends to increase and then decrease as $k$ approaches $k_{max}$, highlighting deficiencies in model performance in accurately classifying fine-grained categories. Importantly, optimal differentiation was not achieved at $k_{max}$, further displaying limitations in distinguishing between closely related categories among Large Vision Language Models. 

\noindent \textbf{Descriptions Generated by In-Context Learning.} In our experiments, we also explored several configurations to enhance the distinctiveness of fine-grained visual descriptions (FGVDs), leveraging the capabilities of In-Context Learning(ICL) rather than solely relying on a zero-shot approach. We utilized methods such as Random Sampling (RS), Similarity-based Image-Image Retrieval (SIIR)~\cite{yang2023exploring}, and Similarity-based Text-Text Retrieval (STTR), along with high-quality text generation from GPT-4, to examine the outcomes in terms of distinctiveness. Specifically, for the STTR approach, we generated a textual case using a zero-shot framework and then identified similar descriptions within the supporting set. We construct In-Context Examples (ICEs) based on these similarities, which assist the model in generating fine-grained visual descriptions, other detailed selection methods will be demonstrated in the Appendix.
As shown in Table~\ref{icl}, the use of In-Context Learning (ICL) for auxiliary generation in methods such as RS, SIIR, and STTR resulted in poorer outcomes. Our analysis suggests that this decline in performance is attributable to the use of model-generated, fine-grained visual descriptions that were of low quality. Employing ICL with subpar descriptions initiates a detrimental cycle that degrades overall results. To address this, we incorporated GPT-4~\cite{achiam2023gpt} to produce high-quality FGVDs, which significantly improved results across both datasets when assisting in LVLM generation. Our results demonstrate that employing high-quality, long-form, fine-grained visual descriptions not only boosts performance but also mitigates issues of model hallucination, further discussed in Section \ref{subsec:quality}.

\subsection{Fidelity Evaluation}
\label{subsec:fidelity}

This section focuses on assessing the fidelity of fine-grained visual descriptions produced by Large Vision-Language Models. To this end, LVLMs are prompted with a single question without demonstrations, requiring a comprehensive understanding of the visual content of the image. This forces the models to articulate visual information in a textual format. The uniform prompt used is 

\textit{"What are the main elements in this image, and how do they interact or relate to each other?"}. 

\begin{table}[t]
\centering
\caption{Image-text fidelity results across five datasets utilizing the CLIP-S metric (CLIP embedding similarity).}
\label{tab:i-t}
    \scalebox{0.98}{
    \begin{tabular}{lccc} 
    \toprule
    \textbf{Dataset} & IDE & MiniGPT4 & OF \\ 
    \midrule
    CUB200           & 24.06 & \textbf{26.63} & 24.81 \\
    Flowers102       & 25.13 & \textbf{27.72} & 25.31 \\
    Stanford Cars    & 25.16 & \textbf{26.51} & 25.22 \\
    Stanford Dogs    & 22.58 & \textbf{27.27} & 25.20 \\
    ImageNet Subset  & 25.12 & \textbf{27.80} & 25.78 \\
    \bottomrule
    \end{tabular}
    }

\end{table}
\begin{figure}[t]
    \centering
    \includegraphics[width=0.96\linewidth]{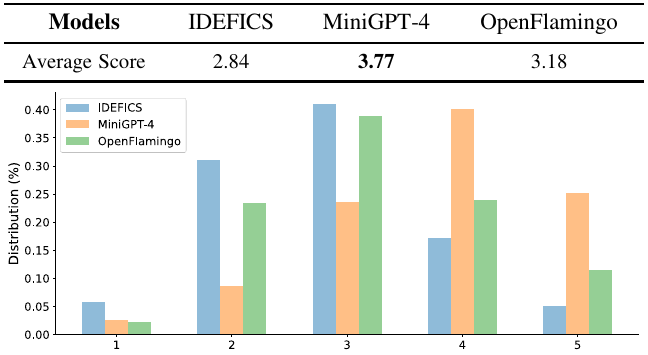}
    \caption{Human evaluation results assessing the fidelity of descriptions generated by LVLMs. Top: Average scores across various models. Bottom: Score distribution from 1 to 5 for each model.}

    \label{fig:human_eval}
\end{figure}
\begin{table}[t]
\centering
\caption{Experimental results for image-reconstructed image fidelity across five datasets utilizing three metrics (\ie, SSIM $\uparrow $, FID $\downarrow $, CLIP-S-I (CLIP embedding similarity) $\uparrow $).}
\label{tab:i-i}
\scalebox{0.6}{
\begin{tabular}{lccccccccc} 
\toprule
\multirow{2}{*}{\textbf{ Dataset}} & \multicolumn{3}{c}{SSIM $\uparrow$ }    & \multicolumn{3}{c}{FID $\downarrow $}     & \multicolumn{3}{c}{CLIP-S-I $\uparrow$}   \\ 
\cmidrule(lr){2-4} \cmidrule(lr){5-7} \cmidrule(lr){8-10}
                                   & IDE & MiniGPT4 & OF &  IDE & MiniGPT4 & OF & IDE & MiniGPT4 & OF  \\ 
\cmidrule{1-1}\cmidrule{2-4} \cmidrule{5-7} \cmidrule{8-10}
CUB200                             & 28.25  & \textbf{29.52}    & 28.69    & 90.98     & \textbf{71.28}    & 93.37     & 74.83    & \textbf{75.98}    & 73.64     \\
Flowers102                         & 18.39    & \textbf{20.37}    & 19.33    & 255.57    & \textbf{215.62}   & 318.54    & 75.49    & \textbf{79.15}    & 76.30     \\
Stanford Cars                      & 16.22    & \textbf{17.38}    & 16.80    & 63.69     & \textbf{49.61}    & 77.38     & 73.35    & \textbf{75.87}    & 73.69     \\
Stanford Dogs                      & 16.17    & \textbf{17.16}    & 16.96    & 329.75    & \textbf{218.29}   & 252.75    & 67.87    & \textbf{73.45}    & 71.55     \\
ImageNet Subset                    & 17.03    & \textbf{17.55}    & 17.47    & 78.20     & \textbf{66.84}    & 64.14     & 71.33    & \textbf{73.54}    & 71.67     \\
\bottomrule
\end{tabular}
}
\end{table}
\noindent\textbf{MiniGPT-4 Exhibits Higher Fidelity.} We began our analysis by utilizing the CLIP Embedding Similarity (CLIP-S) metric, which measures the semantic similarity between the generated FGVDs (Fine-Grained Visual Descriptions) and the original images. As indicated in Table~\ref{tab:i-t}, FGVDs produced by MiniGPT-4 exhibit higher fidelity compared to those generated by the other two models. Despite the limitations imposed by modal discrepancies within the CLIP embedding space, as noted in previous studies~\cite{liang2022mind}, the descriptions generated by MiniGPT-4 consistently achieved the highest scores across multiple datasets.

To complement our quantitative findings and enhance the comprehensiveness of our fidelity assessment, we integrated human evaluation into our experiment. This method assesses the alignment between the generated FGVDs and their corresponding original images, assigning scores from 1 to 5 to denote increasing levels of fidelity. The evaluation also considers potential issues such as hallucinations and the introduction of external knowledge. For this purpose, we randomly selected 20 FGVDs across five datasets. The results of this human evaluation are displayed in Figure~\ref{fig:human_eval}, where MiniGPT-4 achieved the highest scores. Furthermore, the number of high-fidelity FGVDs produced by MiniGPT-4 significantly surpassed those generated by the other models, demonstrating its superior performance in maintaining textual-visual alignment.
% ~\cy{Provide the possible reason why MINI is better from architecture, data, or other perspectives}.

\begin{figure*}[t]
    \centering
    \includegraphics[width=1\linewidth]{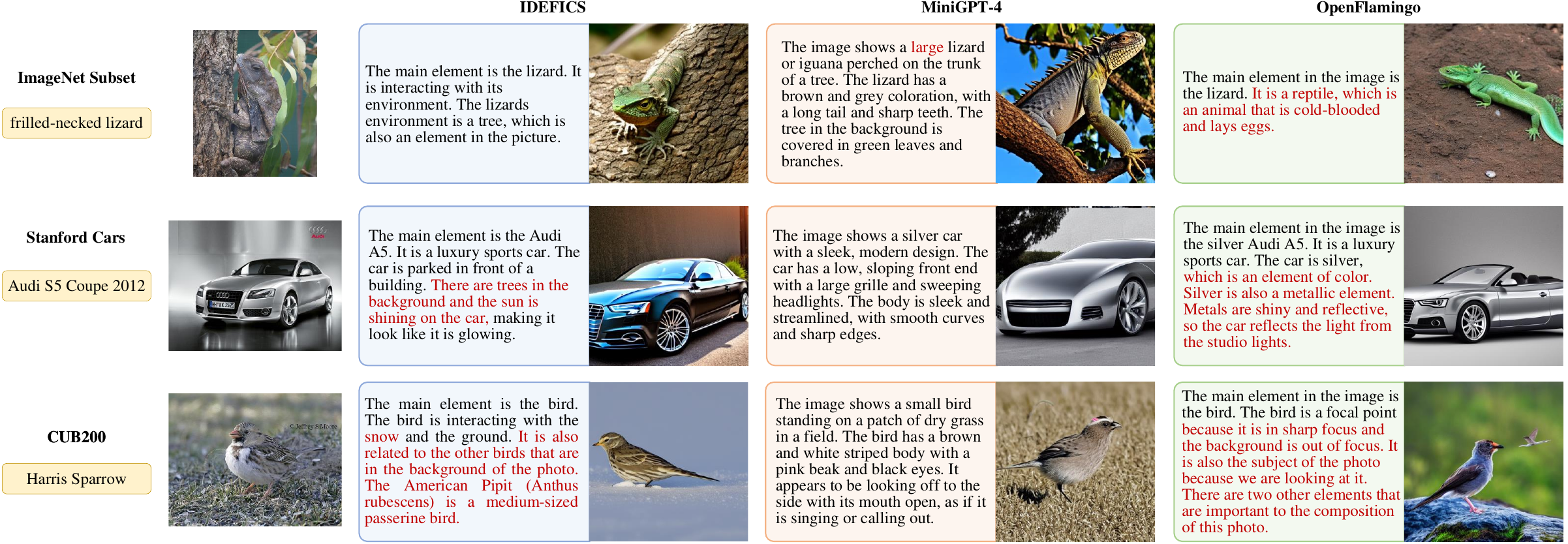}
    \caption{Qualitative examples of fine-grained visual descriptions (FGVDs) generated by three Large Vision-Language Models (LVLMs), alongside their corresponding reconstructed images. Hallucinations in the generated FGVDs are highlighted in red to indicate discrepancies between the generated descriptions and the actual image content.}
    \label{fig:quality}
\end{figure*}

\noindent\textbf{Image Reconstruction Amplifies Discrepancies in LVLM Outputs.}
We further scrutinized the fidelity of Large Vision Language Models (LVLMs) in preserving visual information during generation. Utilizing Stable Diffusion, we transformed the generated Fine-Grained Visual Descriptions (FGVDs) back into reconstructed images for a thorough analysis. To assess fidelity, we employed metrics such as Structural Similarity Index (SSIM)~\cite{wang2004image} and CLIP Embedding Similarity (CLIP-S-I), comparing the original and reconstructed images. Additionally, the Fréchet Inception Distance (FID)~\cite{heusel2017gans} was used to evaluate the distributional distance between the original and reconstructed images, offering insights into how effectively the text, serving as an intermediary, preserves visual information.
Table~\ref{tab:i-i} reaffirmed the superior performance of MiniGPT-4 in this complex task. Throughout the process of image reconstruction, text acts as a crucial intermediary, adeptly preserving information that remains faithful to the original image. However, any discrepancies or elements unfaithful to the original were significantly exacerbated in the reconstructed images. This not only augmented our evaluation but also illuminated the limitations and capabilities of LVLMs. As shown in Fig.~\ref{fig:quality}, the FGVDs regenerated by Open-Flamingo and IDEFICS exhibited notable dissimilarities to the original images, highlighting the challenges of producing faithful visual representations.
\section{Qualitative Results}

% ~\cy{insights: Generated images can be used as an intermediary to measure text-image similarity. More explanation on the correlation of the evaluation results between IT fidelity and I-RI fidelity. Explain the advantages of I-RI from analyzing the evaluation results.}
% Upon closer examination, all three models struggle with the Stanford Dogs dataset, underscoring the challenge it poses for LVLMs in generating high-fidelity visual descriptions.

\begin{figure}[t]
    \centering
    \includegraphics[width=1.0\linewidth]{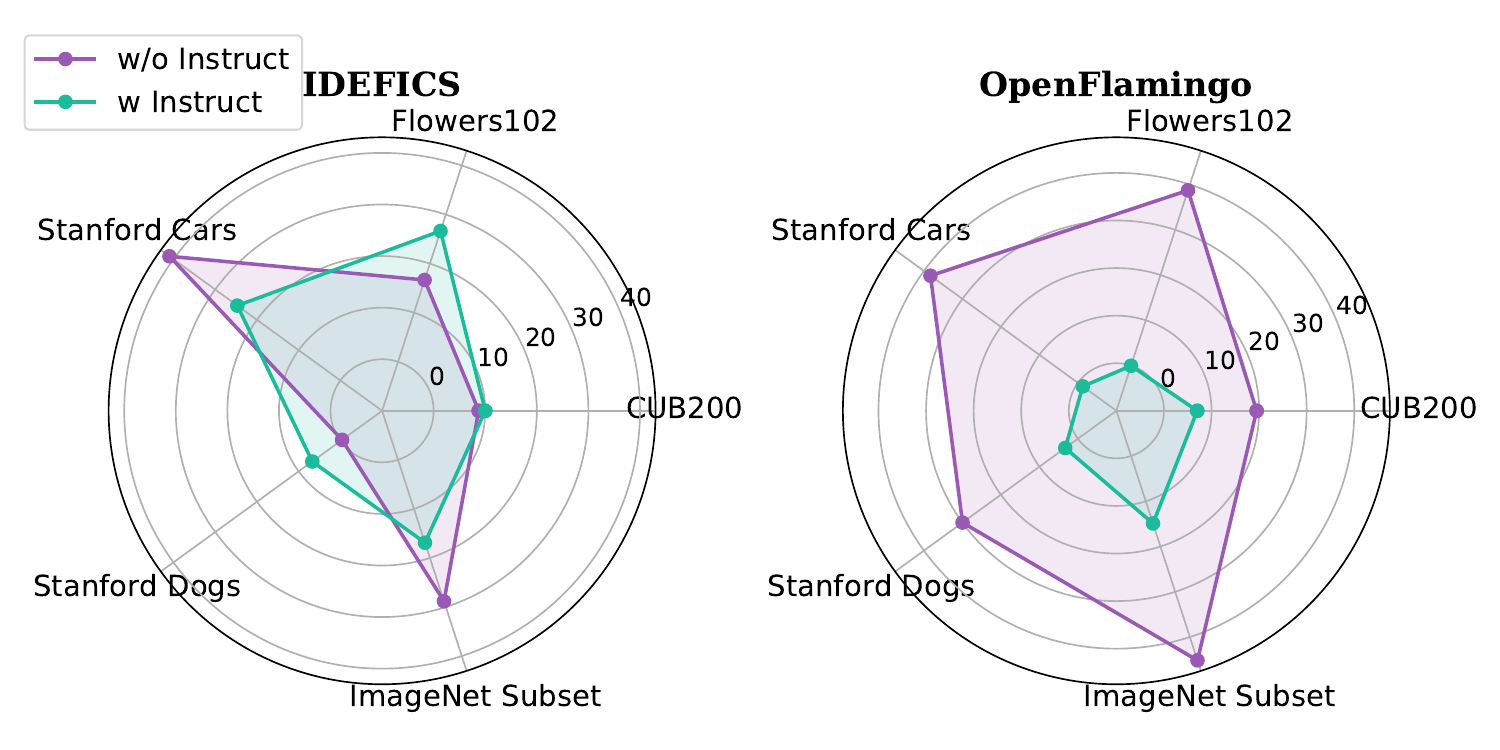}
    \caption{Comparative results with and without instructional guidance for description.}
    \label{fig:enter-label}
\end{figure}

\begin{figure}[t]
    \centering
    \includegraphics[width=1.0\linewidth]{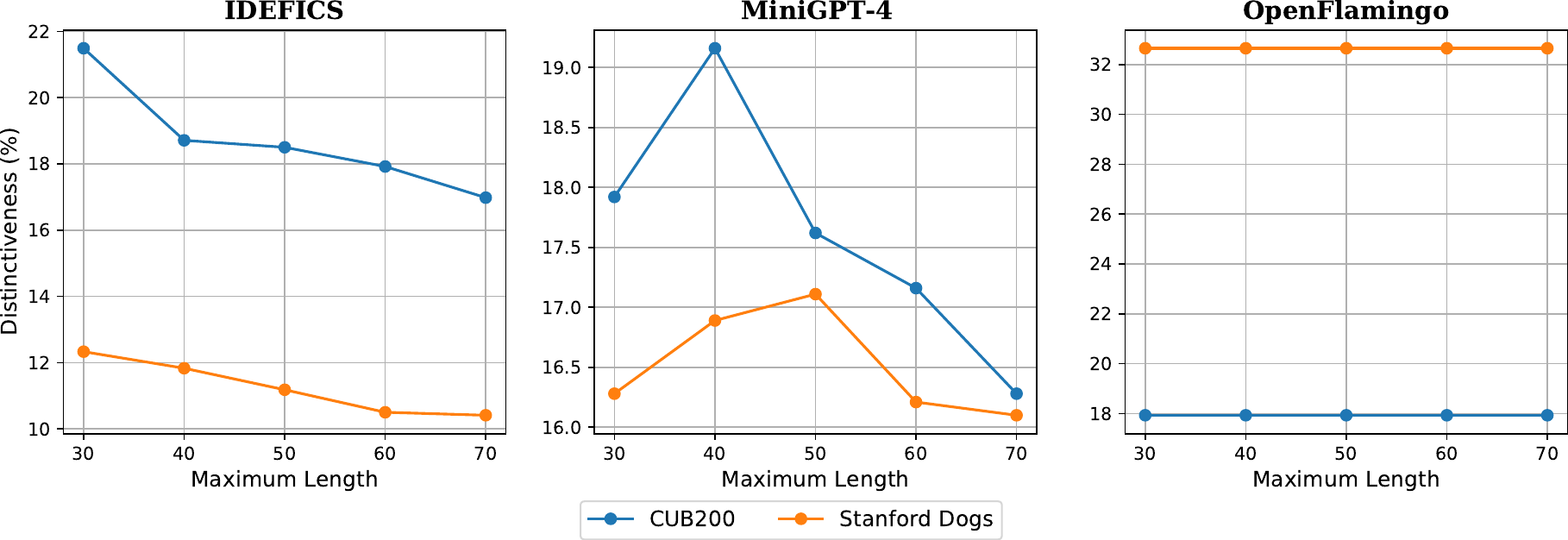}
    \caption{Distinctiveness of LVLMs-Generated fine-grained text at different lengths.}
    \label{fig:LENGTH}
\end{figure}
\subsection{Hallucination Analysis}
\label{subsec:quality}

\begin{table}
\centering
\caption{Using different configurations in In-Context Learning to help LVLMs generate. (l30 means length of 30)}
\label{icl}
\scalebox{0.56}{
\begin{tabular}{c|ccccc|ccccc} 
\hline
\multicolumn{1}{c}{\multirow{2}{*}{Methods And Datasets}}                               & \multicolumn{5}{c}{CUB200}                                                                                                                                              & \multicolumn{5}{c}{Stanford Dogs}                                                                                                                                               \\
\multicolumn{1}{c}{}                                                                    & \multicolumn{1}{l}{0-shot} & \multicolumn{1}{l}{1-shot}         & \multicolumn{1}{l}{2-shot} & \multicolumn{1}{l}{3-shot}         & \multicolumn{1}{l}{4-shot}          & \multicolumn{1}{l}{0-shot} & \multicolumn{1}{l}{1-shot}         & \multicolumn{1}{l}{2-shot}         & \multicolumn{1}{l}{3-shot}         & \multicolumn{1}{l}{4-shot}          \\ 
\hline
RS (l30)                                                                                & 21.49                      & 8.78                               & 8.54                       & 9.51                               & 9.22                                & 12.33                      & 7.27                               & 7.87                               & 8.61                               & 9.03                                \\
RS (l70)                                                                                & 16.98                      & 10.11                              & 12.08                      & 12.5                               & 10.96                               & 10.41                      & 6.28                               & 8.29                               & 9.36                               & 9.79                                \\
SIIR (l30)                                                                              & 21.49                      & 6.71                               & 7.68                       & 9.18                               & 7.8                                 & 12.33                      & 6.08                               & 5.78                               & 5.93                               & 6.32                                \\
SIIR (l70)                                                                              & 16.98                      & 6.17                               & 8.7                        & 8.66                               & 8.77                                & 10.41                      & 5.22                               & 4.9                                & 5.16                               & 5.54                                \\
STTR (l30)                                                                              & 21.49                      & 6.45                               & 7.56                       & 8.03                               & 8.20                                & 12.33                      & 5.05                               & 6.85                               & 6.99                               & 7.07                                \\
STTR (l70)                                                                              & 16.98                      & 7.65                               & 9.20                       & 9.22                               & 9.29                                & 10.41                      & 5.02                               & 5.87                               & 6.47                               & 7.02                                \\
\begin{tabular}[c]{@{}c@{}}\textbf{GPT-4 Generated}\\\textbf{~Description}\end{tabular} & 16.98                      & \multicolumn{1}{l}{\textbf{18.21}} & \textbf{19.00}             & \multicolumn{1}{l}{\textbf{19.02}} & \multicolumn{1}{l|}{\textbf{22.87}} & 10.41                      & \multicolumn{1}{l}{\textbf{16.36}} & \multicolumn{1}{l}{\textbf{16.71}} & \multicolumn{1}{l}{\textbf{17.96}} & \multicolumn{1}{l}{\textbf{19.64}}  \\
\hline
\end{tabular}
}
\end{table}

After thoroughly checking the quality of the texts generated by the LVLM, we found that all three models have more or less illusory problems. The descriptions will be inconsistent with the original images, which is one of the reasons why we introduced Fidelity, to see whether the LVLMs can be discriminative and at the same time have fidelity from another perspective.

\noindent \textbf{LVLMs Exhibits Non-Relevant Information.} In our distinctiveness experiment, we observed that MiniGPT-4 consistently produces accurate descriptions aligned with the images. In contrast, IDEFICS and Open-Flamingo often generate text that, while clear, is unrelated to the image content. They tend to prematurely specify the label first at the beginning of sentences, leading to content that does not align with the visual information. This approach results in subsequent descriptions becoming completely disordered. For example, the description corresponding to the category \textit{"Otterhound"} was \textit{"Airedale Terrier. \textbf{Photo Credit: Wikimedia Commons.}"} This was particularly evident after modifying the prompts to enhance visual detail representation by adding \textit{"It has"} to the descriptions, which helped steer the LVLMs towards more visually representative outputs. As illustrated in Fig. \ref{fig:enter-label}, the comparative performance before and after this guidance highlights the relative weakness of Open-Flamingo in processing visual information, explaining its initial superiority on the Stanford Dogs dataset. Overall, MiniGPT-4 produces more stable and higher-quality descriptions than IDEFICS and Open-Flamingo, attributable to its simplified architecture, targeted alignment dataset, and refined training approach, which collectively sharpen accuracy and reduce hallucinatory content.

\noindent \textbf{Increased Length Leads to Loss of Focus and Hallucinations in LVLMs.} Upon further analysis, we observed an increase in the probability of nonsensical output as sentence length extended. We evaluated three models—Open-Flamingo, IDEFICS, and MiniGPT-4, using maximum lengths ranging from 30 to 70 while keeping the length penalty parameter constant. Our findings in Fig. \ref{fig:LENGTH} indicate that Open-Flamingo is insensitive to these length changes, typically generating shorter texts. In contrast, IDEFICS and MiniGPT-4 are more likely to produce longer texts, a tendency we attribute to differences in the training datasets utilized~\cite{laurenccon2024obelics}. Importantly, while both IDEFICS and MiniGPT-4 showed decreased distinctiveness at the maximum length of 70, the performance trend differed: IDEFICS exhibited a consistent decline in distinctiveness as text length increased, whereas MiniGPT-4 initially showed an improvement in distinctiveness, which peaked before eventually declining as lengths approached 70. The decrease in distinctiveness with increasing text length, leads to a higher likelihood of producing content that is irrelevant to the image, for example, the generated description often contains \textit{"\textbf{It is looking at the camera}", "\textbf{There is a blue sky}"} at the end of the description, however, this is not accurate and irrelevant with the visual features. We believe these results reveal a limitation of LVLMs: their diminishing ability to stay visually coherent as text lengthens. This issue suggests a need for training methods that better maintain context over longer outputs. We hypothesize that rigorous post-processing and manual verification of the training data are critical to achieving this.

\section{Conclusion and Limitations}
\label{sec:conclusion}

In this paper, we focused on exploring the distinctiveness and fidelity of textual descriptions generated by Large Vision-Language Models (LVLMs) using our proposed TRAC method coupled with various analytical techniques. Our analysis revealed that MiniGPT-4 excels in generating fine-grained descriptions, marking the first evaluation of a distinctive and faithful generation of LVLMs in this domain. This work enriches multimodal language model research and identifies critical areas for improvement, especially in addressing hallucination issues inherent to these models. 
However, relying on existing text encoders for feature extraction may limit our study, as the CLIP embeddings, along with other current embedding techniques, have limited capability in distinguishing fine-grained details, impacting the evaluation score. Additionally, our reliance on a single Stable Diffusion model may also limit the generalization of our findings; future studies could benefit from testing across various types of Stable Diffusion models to validate and refine our results and we plan to enhance our methods for better granularity in distinction and explore new classification strategies for LVLMs using our findings.

% \appendix
% \input{sec/X_suppl}

% \section{Acknowledgments}
% Identification of funding sources and other support, and thanks to
% individuals and groups that assisted in the research and the
% preparation of the work should be included in an acknowledgment
% section, which is placed just before the reference section in your
% document.

% This section has a special environment:
% \begin{verbatim}
%   \begin{acks}
%   ...
%   \end{acks}
% \end{verbatim}
% so that the information contained therein can be more easily collected
% during the article metadata extraction phase, and to ensure
% consistency in the spelling of the section heading.

% Authors should not prepare this section as a numbered or unnumbered {\verb|\section|}; please use the ``{\verb|acks|}'' environment.

% \section{Appendices}

% If your work needs an appendix, add it before the
% ``\verb|\end{document}|'' command at the conclusion of your source
% document.

% Start the appendix with the ``\verb|appendix|'' command:
% \begin{verbatim}
%   \appendix
% \end{verbatim}
% and note that in the appendix, sections are lettered, not
% numbered. This document has two appendices, demonstrating the section
% and subsection identification method.

%%
%% The acknowledgments section is defined using the "acks" environment
%% (and NOT an unnumbered section). This ensures the proper
%% identification of the section in the article metadata, and the
%% consistent spelling of the heading.
% \begin{acks}
% To Robert, for the bagels and explaining CMYK and color spaces.
% \end{acks}

%%
%% The next two lines define the bibliography style to be used, and
%% the bibliography file.
% \newpage
\bibliographystyle{ACM-Reference-Format}
\bibliography{main}
\appendix
\setcounter{page}{1}

\renewcommand{\thesection}{\Alph{section}}
\renewcommand{\thesubsection}{\thesection.\arabic{subsection}}
% \section{Few-shot Learning to recognize new classes}
\section{Datasets}
To evaluate the distinctiveness and fidelity of fine-grained descriptions generated by large vision-language models (LVLMs), we selected specific fine-grained image datasets. These include CUB-200, Stanford Dogs, Stanford Cars, and Oxford 102 Flowers. To assess the performance of a more generic objects dataset, we curated a subset from ImageNet, consisting of 10 samples per class for the training set and 5 samples per class for the testing set. Detailed descriptions of these datasets are provided in Table~\ref{tab:datasets}.

\begin{table}[h]
\caption{Dataset Information}
\begin{tabular}{lccc}
\toprule
\textbf{Dataset} & \textbf{Classes} & \textbf{Support Set} & \textbf{Test Set} \\
\midrule
CUB200 & 200 & 5994 & 5794 \\
Flowers102 & 102 & 4093 & 2463 \\
Stanford Cars & 196 & 8144 & 8041 \\
Stanford Dogs & 120 & 12000 & 8580 \\
ImageNet Subset & 1000 & 10000 & 5000 \\
\bottomrule
\end{tabular}

\label{tab:datasets}
\end{table}
\section{Comparison}
\label{app:comp}
To avoid the impact of the differences between different text encoders on our \textbf{TRAC} method, we tested the text encoder of CLIP and Sentence-BERT, and we even utilized the TF-IDF statistic method to observe the results. As shown in Table \ref{tab:com}, after careful consideration, we used the embeddings generated by the text encoder of CLIP for the experiments in this paper.
\begin{table*}[btp]
\centering
\caption{Distinctiveness results obtained using different methods or models for retrieval.}
\label{tab:com}
\scalebox{0.82}{
\begin{tabular}{cccccccccccccccc} 
\hline
                                & \multicolumn{5}{c}{CLIP-Embeddings}   & \multicolumn{5}{c}{SENTENCE-BERT-Embeddings} & \multicolumn{5}{c}{TF-IDF}             \\ 
\hline
\multirow{2}{*}{IDEFICS}      & CUB   & DOG   & CAR   & FLOWER     & IMAGENET     & CUB   & DOG   & CAR   & FLOWER     & IMAGENET            & CUB   & DOG   & CAR   & FLOWER     & IMAGENET      \\
                                & 21.49 & 12.33 & 52.61 & 34.63 & 33.66 & 22.07 & 13.00 & 51.89 & 28.83 & 34.90        & 19.59 & 12.03 & 49.02 & 31.91 & 27.46  \\
\multirow{2}{*}{OpenFlamingo} & CUB   & DOG   & CAR   & FLOWER     & IMAGENET     & CUB   & DOG   & CAR   & FLOWER     & IMAGENET            & CUB   & DOG   & CAR   & FLOWER     & IMAGENET      \\
                                & 27.20 & 32.65 & 45.60 & 50.83 & 46.34 & 17.47 & 31.48 & 48.25 & 46.62 & 41.30        & 15.27 & 29.63 & 47.82 & 46.49 & 39.50  \\
\multirow{2}{*}{MiniGPT-4}       & CUB   & DOG   & CAR   & FLOWER     & IMAGENET     & CUB   & DOG   & CAR   & FLOWER     & IMAGENET            & CUB   & DOG   & CAR   & FLOWER     & IMAGENET      \\
                                & 19.16 & 17.11 & 47.00 & 52.78 & 39.94 & 21.83 & 14.87 & 42.46 & 49.21 & 41.36        & 16.76 & 11.45 & 43.12 & 42.79 & 35.72  \\
\hline
\end{tabular}
}
\end{table*}

\section{Stanford Dog Poor Generation}
% We found that in the Stanford Dogs dataset, the distinctiveness generated by LVLMs works well in OpenFlamingo. After detailed examination, we found that it is because the model tends to generate some strange statements and extra knowledge as shown in Fig. \ref{fig:demo}, and OpenFlamingo tends to generate shorter statements, so the scores obtained can be relatively high. Meanwhile, we can find that MiniGPT-4 answered most closely to the visual information of the images by comparing them with the corresponding images.

\begin{figure}[hbtp]
    \centering
    \includegraphics[width=1.2\linewidth]{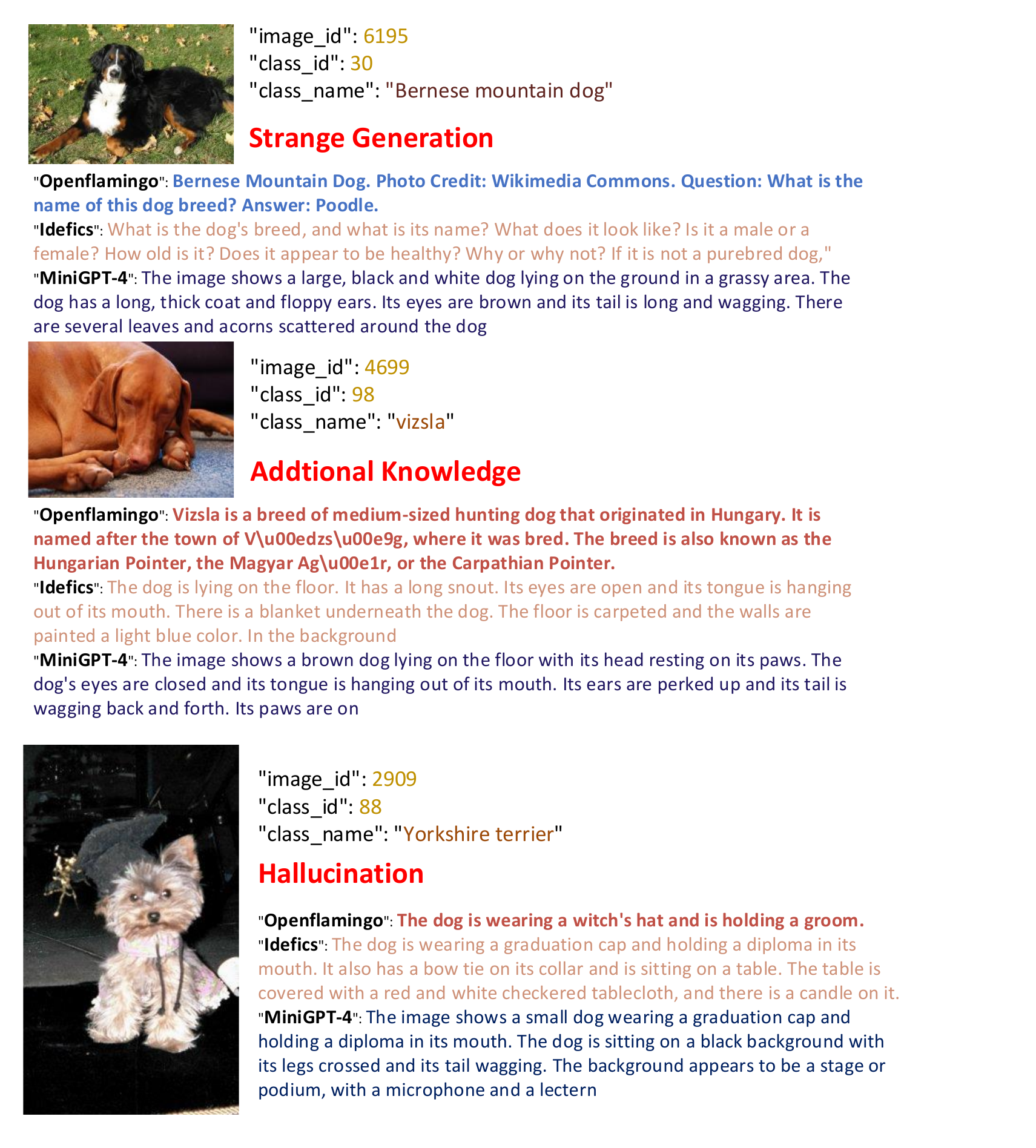}
    \caption{Poor generation of Stanford Dog among 3 LVLMs.}
    \label{fig:demo}
\end{figure}

Our analysis of the Stanford Dogs dataset revealed that while Large Vision-Language Models (LVLMs) generally perform well in generating distinctive descriptions, especially in OpenFlamingo, certain issues persist. Detailed examinations indicate that the model occasionally produces strange statements and incorporates external information, as illustrated in Fig. \ref{fig:demo}. Notably, OpenFlamingo tends to generate short descriptions, which may artificially inflate its performance scores. In contrast, MiniGPT-4 consistently generates descriptions that align more closely with the visual content of the images, as demonstrated through side-by-side comparisons with the corresponding images.

% \begin{figure*}[h]
%     \centering
%     \includegraphics[width=1\linewidth]{fig/qua}
%     \caption{Qualitative Results.}
%     \label{fig:quality}
% \end{figure*}

\section{Human Evaluation for the Fidelity}
To more accurately assess the fidelity of fine-grained visual descriptions (FGVDs) generated by three different Large Vision-Language Models (LVLMs), we have implemented a human evaluation protocol. Considering the complexity involved in the evaluation process, we randomly selected 20 sets of data from five datasets used in our study as subjects for this human evaluation. Furthermore, we developed a 5-point evaluation rubric that accounts for both the presence of hallucinatory elements and the introduction of external knowledge within the descriptions. This rubric is designed to rigorously evaluate the quality of the FGVDs generated by LVLMs, ensuring a comprehensive and unbiased assessment of their performance in creating visually congruent textual descriptions. The criteria for scoring are shown in Table~\ref{tab:human_evaluation}.
\begin{table*}[hbtp]
\centering
\caption{Human Evaluation Rubric for Fidelity of Fine-Grained Visual Descriptions}
\label{tab:human_evaluation}
\begin{tabular}{cp{12cm}}
\toprule
\textbf{Score} & \textbf{Description} \\
\midrule
\textbf{1 }             & \textbf{Completely Unfaithful:} There is no relation between the description and the original image. The elements mentioned do not exist in the image. \\
\textbf{2}              & \textbf{Almost Unfaithful:} There are minimal similarities between the description and the original image. The description largely integrates external knowledge. Most contents either do not correspond to or significantly deviate from the actual image. \\
\textbf{3}              & \textbf{Partially Faithful:} The description somewhat matches the original image with a fair proportion of accuracy, albeit with some external knowledge text. Some content still mismatches or omits key elements of the image. \\
\textbf{4}              & \textbf{Mostly Faithful:} Most of the description is consistent with the original image, with minor discrepancies or omissions. \\
\textbf{5}              & \textbf{Completely Faithful:} The description perfectly aligns with the original image with sufficient detail. All described elements can be identified in the image without any omissions or inaccuracies. \\
\bottomrule
\end{tabular}
\end{table*}

\end{document}